\documentclass[preprint]{elsarticle}
\usepackage{graphicx}
\usepackage{listings}
\usepackage{multirow}
\usepackage[pdftex]{hyperref}
\begin{document}

\begin{frontmatter}
\title{Merging of neural networks}
%
%

\author[1]{Martin Pašen}
\author[1]{Vladimír Boža}

\affiliation[1]{organization={Faculty of Mathematics, Physics and Informatics, Comenius University
in Bratislava}, country={Slovakia}}

%
%
%
%
\begin{abstract}

We propose a simple scheme for merging two neural networks trained with different starting initialization into a single one with the same size as the original ones. We do this by carefully selecting channels from each input network. Our procedure might be used as a finalization step after one tries multiple starting seeds to avoid an unlucky one.
We also show that training two networks and merging them leads to better performance than training a single network for an extended period of time.
\footnote{
Availability: \href{https://github.com/fmfi-compbio/neural-network-merging}{\detokenize{https://github.com/fmfi-compbio/neural-network-merging}}
}
\end{abstract}

\end{frontmatter}
\section{{Introduction}}

Typical neural network training starts with random initialization and is trained until reaching convergence in some local optima.
  The final result is quite sensitive to the starting random seed as reported in \cite{picard2021torch,wightman2021resnet}, who observed 0.5\% difference in accuracy between worst and best seed on Imagenet dataset and 1.8\% difference on CIFAR-10 dataset.
  Thus, one might need to run an experiment several times to avoid hitting the unlucky seed.
  The final selected network is just the one with the best validation accuracy.

We believe that discrepancy between starting seeds performance can be explained by selecting slightly different features in hidden layers in each initialization.
  One might ask a question, can we somehow select better features for network training?
One approach is to train a bigger network and then select the most important channels via channel
pruning  \cite{molchanov2016pruning,molchanov2019importance,luo2017thinet,yu2018nisp}.
Training a big network, which is subsequently pruned, might be in many cases prohibitive, since increasing network width by a factor of two results in a four times increase in FLOPs and also might require a change in some hyperparameters (e.g.
  regularization, learning rate).

  \begin{figure}[h!]
    \centering
  
    \includegraphics[width=0.7\textwidth,clip,trim=0 2cm 0 0]{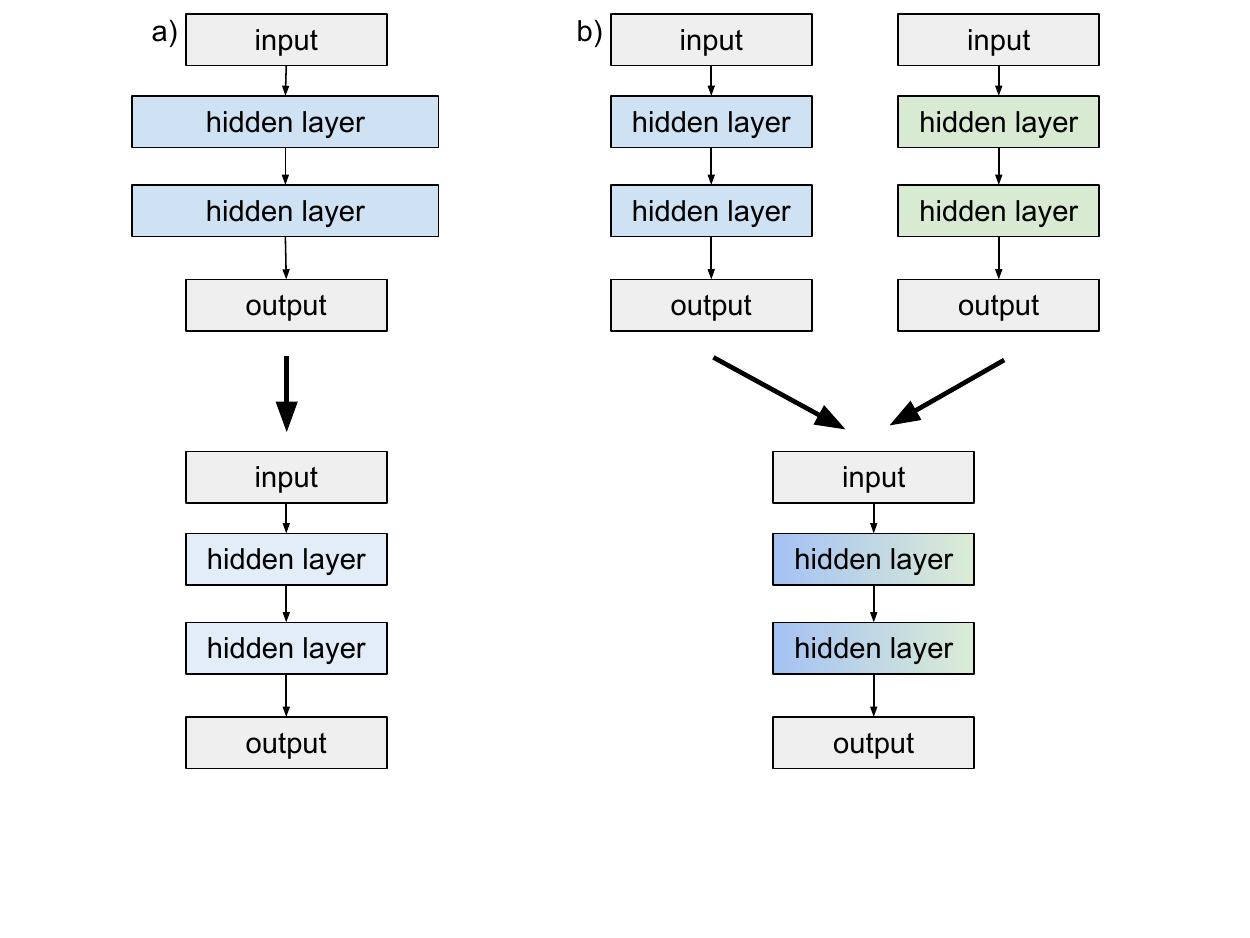}
    
    \caption{
       \label{fig:prune_vs_merge} Comparison between a) training a bigger network and then
      pruning and b) training two separate networks and then merging them together.
          Width of rectangles denotes the number of channels in the layer.}
    
  \end{figure}

  \begin{figure}
  \centering
  \includegraphics[width=1.0\textwidth,clip,trim=3cm 0 3cm 0]{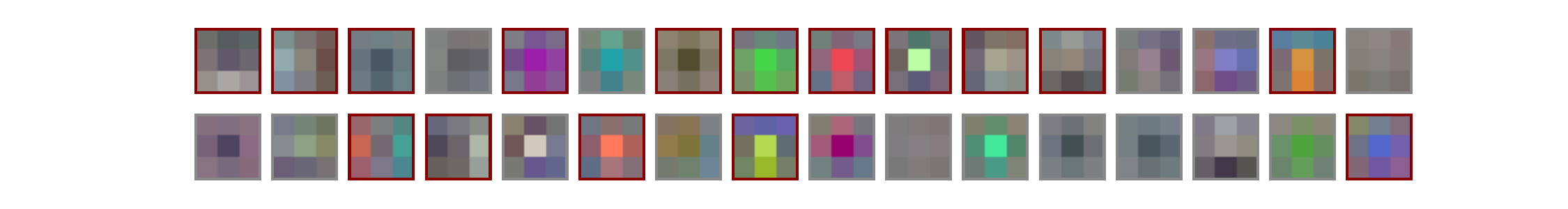}
  \caption{
  \label{fig:filters} Set of filters in the first layer of two ResNet20 networks trained on
  CIFAR-100 dataset with different starting seed. Each row shows filters from one network. 
  Selected filters for merged network are marked with red outline.
  }
  \end{figure}
 
Here, we propose an alternative approach demonstrated in Fig.
  \ref{fig:prune_vs_merge} .
  Instead of training a bigger network and pruning it, we will train two same sized networks and merge them together into one.
  The idea is that each training run would fall into different local optima and thus have different
  sets of filters in each layer as shown in Fig.~\ref{fig:filters}.
  We then can select a better set of filters than in original networks and achieve better accuracy.

In summary, in this paper:
\begin{itemize}

  \item We propose a procedure for merging two networks with the same architecture into one with
  same architecture as original ones.
  \item We demonstrate that our procedure produces a network with better performance than the best of original ones. 
    On top of that, we also show that the resulting network is better than the same network trained for an extended number of epochs (matching the whole training budget for the merged network).
  
\end{itemize}

\subsection{\textbf{Related work}}

There are multiple approaches, which try to improve accuracy/size tradeoff for neural networks without the need for specialized sparse computation (such as in case of weight pruning).
Most notable one is \textbf{channel pruning}
\cite{molchanov2016pruning,molchanov2019importance,luo2017thinet,yu2018nisp}.
  Here we first train a bigger network and then select the most important channels in each layer.
  Selection process usually involves assigning some score to each channel and then removing channels with the lowest score.

Another approach is \textbf{knowledge distillation }\cite{hinton2015distilling}.
  This involves first training a bigger network (teacher) and then using its outputs as targets for a smaller network (student).
  It is hypothesized that by using larger network outputs, the smaller network can also learn hidden aspects of data, which are not visible in the dataset labels.
  However, it was shown that successful knowledge distillation requires training for a huge number of epochs (i.e.
  1200) \cite{beyer2021knowledge}.
  A slight twist to distillation was applied in \cite{nath2020better} where bigger and smaller networks were cotrained together.

One can also use auxiliary losses to reduce redundancy and increase diversity between various places in the neural network \cite{chen2022the}.

\section{{Methods}}

Here, we describe our training and merging procedure.
We will denote two networks, which will be merged as \textbf{teachers} and the resulting network as a \textbf{student}.

Our training strategy is composed of three stages:
\begin{enumerate}

  \item Training of two teachers
  
  \item \textbf{Merging procedure}, i.e.
    creating a student, which consists of the following substeps:
  
  \begin{enumerate}
  
    \item Layerwise concatenation of teachers into a big student
    
    \item Learning importance of big student neurons
    
    \item Compression of big student
    
  \end{enumerate}
  
  \item Fine-tuning of the student
  
\end{enumerate}

Training of teachers and fine-tuning of the student is just standard training of a neural network by backpropagation.
  Below, we describe how we derive a student from two teachers.

\subsection{\textbf{Layerwise concatenation of teachers into a big student}}

\begin{figure}[t!]
    \centering
  
    \includegraphics[width=0.7\textwidth,clip,trim=0 5cm 5cm 0]{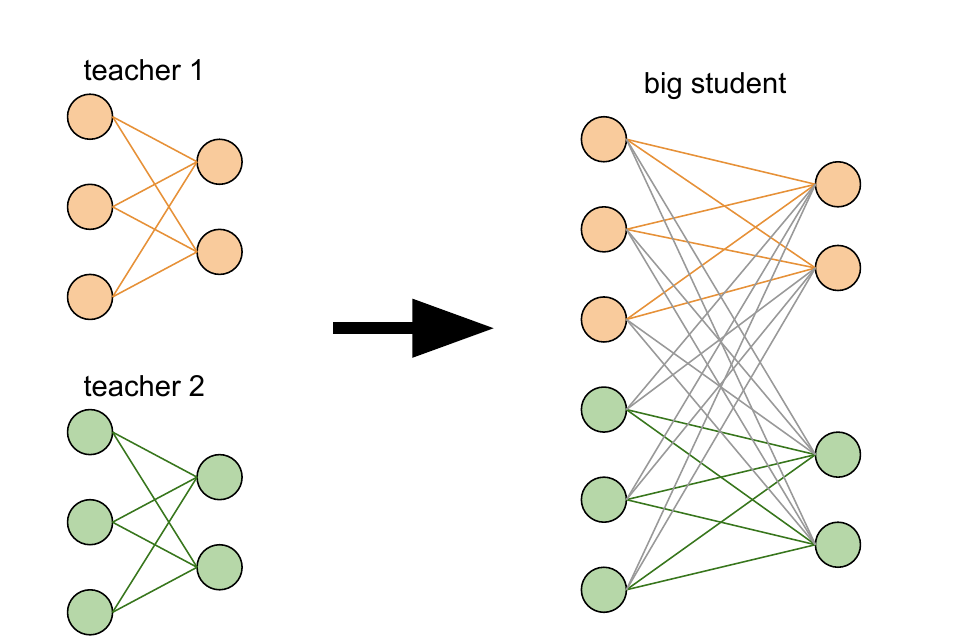}
    
    \caption{
       \label{fig:concatenation} Concatenation of linear layer.
        Orange and green weights are copies of the teacher's weights.
        Gray weights are initialized to zero.
        In the beginning, big student simulates two separate computational flows.
        But during the training, they can be interconnected.
    }
  \end{figure}

First, we create a ``big'' student by layerwise concatenation of teachers.
  The big student simulates the two teachers and averages their predictions in the final layer.
This phase is just network transformation without any training, see Fig.~\ref{fig:concatenation}.
Concatenation of the convolutional layer is done in channel dimension, see Fig.~\ref{fig:concatenation_conv}.
Concatenation of the linear layer is done analogically in the feature dimension.
  We call the model "big student" because it has a doubled width.

\begin{figure}

\begin{lstlisting}[basicstyle=\small]
def merge_conv(conv1: nn.Conv2d, conv2: nn.Conv2d) -> nn.Conv2d:
    in_channels = conv1.in_channels
    out_channels = conv1.out_channels
    conv = nn.Conv2d(in_channels * 2, out_channels * 2,
                     kernel_size=conv1.kernel_size,
                     stride=conv1.stride, padding=conv1.padding,
                     bias=False)
    conv.weight.data *= 0
    conv.weight.data[:out_channels, :in_channels] = \
        conv1.weight.data.detach().clone()
    conv.weight.data[out_channels:, in_channels:] = \
        conv2.weight.data.detach().clone()
    return conv
\end{lstlisting}
\caption{\label{fig:concatenation_conv} Pytorch code for concatenation of a convolutional layer in ResNet.
  Since convolutions are followed by Batch normalization, they do not use biases.
}
\end{figure}


\subsection{\textbf{Learning importance of big student neurons}}

We want the big student to learn to use only half of the neurons in every layer.
So after the removal of unimportant neurons, we will end up with the original architecture.
Besides learning the relevance of neurons, we also want the two computational flows to interconnect.

There are multiple ways to find the most relevant channels.
One can assign scores to individual channels \cite{molchanov2016pruning,molchanov2019importance}, or one can use an auxiliary loss to guide the network to select the most relevant channels.
We have chosen the latter approach, inspired by \cite{voita2019analyzing}.
It leverages the L0 loss presented in \cite{louizos2017learning}.

Let $\ell$ be a linear layer with $k$ input features.
  Let $g_i$ be gate assigned to feature $f_i$.
  Gate can be either opened $g_i =1$ (student is using the feature) or closed $g_i = 0$ (student isn't using the feature).
  Before computing outputs of the layer, we first multiply inputs by gates, i.
  e.
  instead of computing $Wf + b$, we compute $W(f\cdot g) + b$.
To make our model use only half of the features we want $\frac{1}{n}\sum_1^k g_i = \frac{1}{2}$.

The problem with this approach is that $g_i$ is discrete and is not trainable by the gradient descent.
  To overcome this issue, we used stochastic gates and continuous relaxation of $L_0$ norm presented in \cite{louizos2017learning}.
  The stochastic gates contain random variable that has nonzero probability to be 0: $P[g_i=0]>0$, nonzero probability to be 1 $P[g_i=1]>0$, and is continuous on interval $(0,1)$.
  The reparameterization trick makes the distribution of the gates trainable by gradient descent.

To encourage the big student to use only half of the features of the layer, we use an auxiliary loss:
$$L_{half}^\ell = \left(\frac{1}{2} - \frac{1}{k}\sum_1^k P[g_i>0]\right)^2$$

Note that our loss is different from the loss used in \cite{louizos2017learning}.
  Whereas our loss forces the model to have exactly half of the gates opened, their loss pushes the model to use as few gates as possible.

Thus we are optimizing $L = L_{E} + \lambda \sum_\ell L_{half}^\ell$, where $L_{E}$ is error loss
measuring fit on the dataset and new hyperparameter $\lambda$ is proportion of importance of error loss and auxiliary loss.

\begin{figure}[t!]
    \centering
  
    \includegraphics[width=0.6\textwidth]{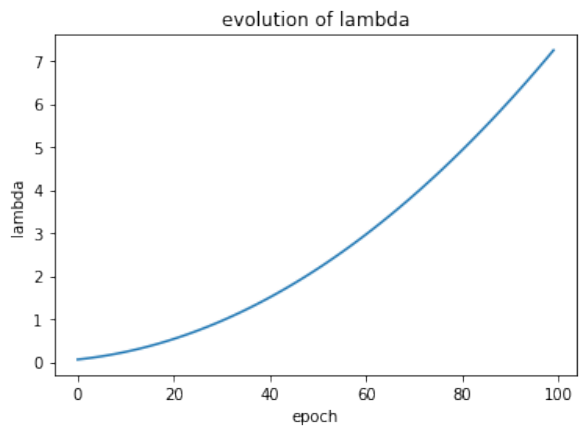}
    
    \caption{
      \label{fig:evolution_of_lambda} Evolution of $\lambda$ during training.
          For the sine problem we have used ${\lambda_{t+1} = \lambda_t  + 0.05* \sqrt{\lambda_t}}$ (where $t$ is the epoch number).
    }
  \end{figure}

Hyperparameter $\lambda$ is sensitive and needs proper tuning.
  At the beginning of the training, it can not be too big, or the student will set every gate to be closed with a probability of 0.5.
  At the end of the training, it can not be too small, or the student will ignore the auxiliary loss in favor of the error loss.
  It will use more than half of the neurons of the layer and will significantly drop performance after the compression.
We found that using the quadratic increase of $\lambda$ during big student's training works sufficiently well, see Fig.
  \ref{fig:evolution_of_lambda}.

\begin{figure}
    \centering
  
    \includegraphics[width=0.7\textwidth]{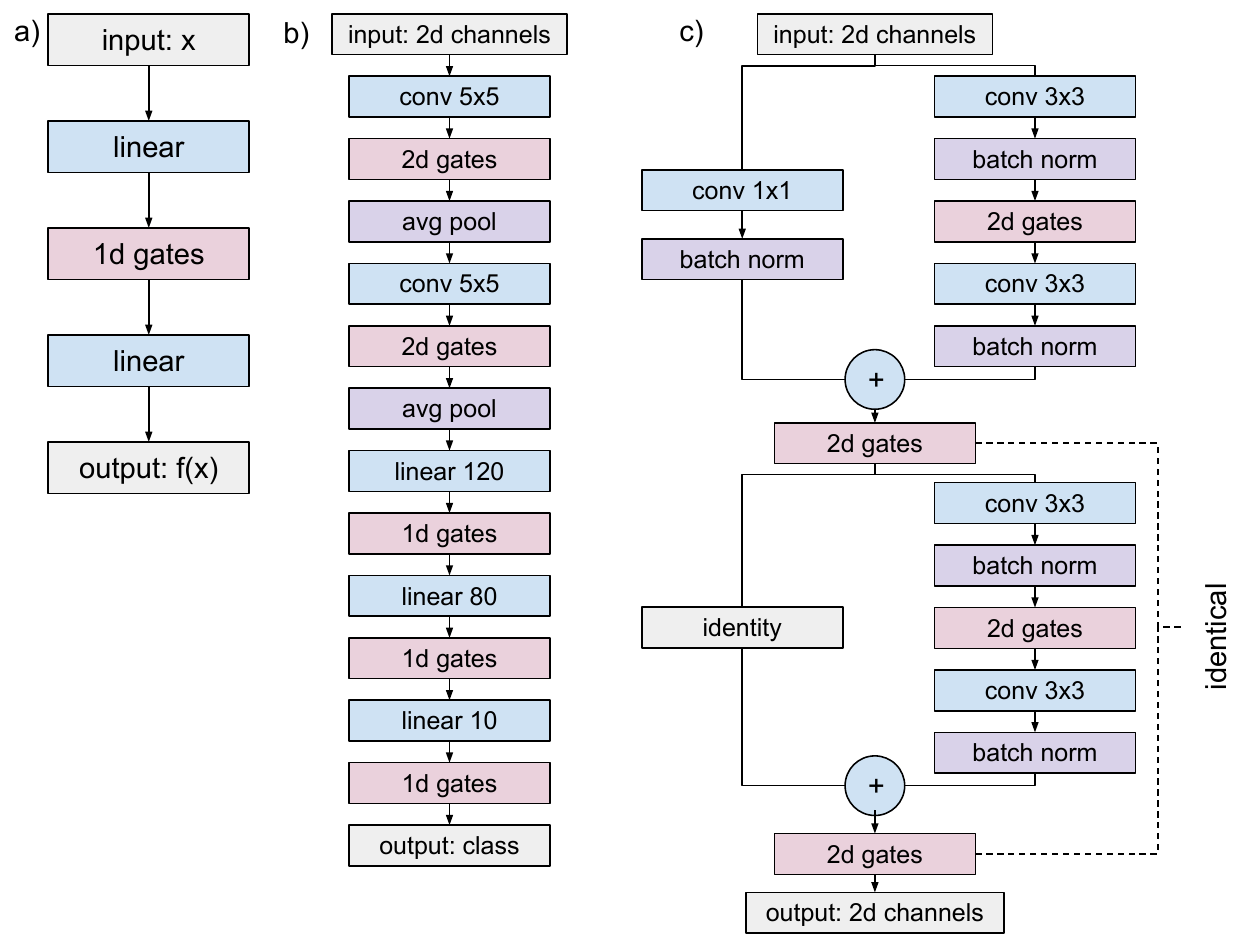}
    
    \caption{ 
\label{fig:network_with_gates} Positions of gate layers a) sine problem b) LeNet c)
two consecutive blocks in ResNet.
  Two of the ResNet gate layers have to be identical.
  If the layers would not be linked and for some channel $i$, the first gate would be closed while the second gate would be opened, the result of the second block, for that channel, would be $0+f(x)$ instead of $x_i + f(x)$ which would defeat the whole purpose of ResNet and skip connections.}
    
  \end{figure}

We have implemented gates in a separate layer.
  We have used two designs of gate layers, one for 2d channels and one for 1d data.
  The position of gate layers is critical.
  For example, if a gate layer is positioned right before the batch norm, its effect (i.
  e.
  multiplying the channel by 0.1) would be countered by the batch norm, see Fig.~\ref{fig:network_with_gates}.

\subsection{\textbf{Compression of big student}}

\begin{figure}[t!]
    \centering
  
    \includegraphics[width=0.7\textwidth,clip,trim=2cm 4.8cm 0 0]{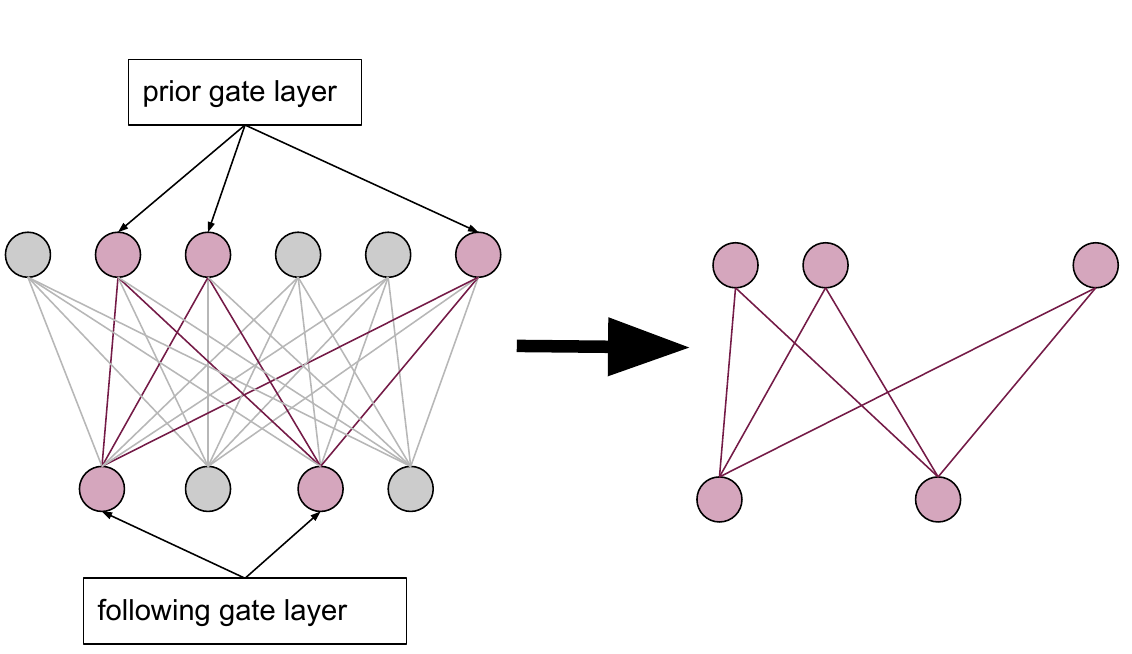}
    
    \caption{
       \label{fig:compression} Compression of the big student. On the left side is a linear layer of a big student.
        Prior and following gate layers decide which neurons are important.
        On the right side is a compressed layer.
        It consists of only the important neurons.
    }
  \end{figure}

After learning of importance is finished, we select half of the most important neurons for every layer.
  Then, we compress each layer by keeping only the selected neurons as is visualized in Fig.~\ref{fig:compression}.

\section{Experimental results}

We compare our merging strategy to generic neural network training on several problems.
First, we test our training strategy on a synthetic problem. We show that our merging strategy can learn better features than typical training. Then we test various architectures on image classification problems.
We show that after merging procedure the resulting network is better than original ones and also
better than training one network for extended amount of time.

\subsection{Training strategies}

We compare our network merging with generic training strategies, which the use same total number of
training epochs.
Except for Imagenet, we are comparing our training strategy with strategies \textit{bo3 model} and \textit{one model}.

In our merging strategy, \textit{student}, we use two-thirds of epochs to train teachers and one-third to train student (one-sixth to find important neurons and one-sixth to fine-tune).

In the \textit{bo3 model} we train three models, each for one-third of epochs, and then we choose the best.

In the final strategy, \textit{one model}, we use all epochs to train one model.

Note that each strategy uses a similar training budget. Also during inference all models
use equivalent amount of resources, since they have exactly the same architecture. 

\subsection{\textbf{Synthetic dataset - Sine problem}}

\begin{table*}[t!]
  \centering

    \caption{
 \label{fig:sine_results_tables}
 Summary of experimental results for sine problem. We report square loss for each strategy.
  Strategy \textit{student} (our strategy) uses 2/3 to train two teachers and 1/3 to train student (1/6
  finding important features, 1/6 finetuning).
  Strategy \textit{bo3 model} trains three models and picks the best.
  Strategy \textit{one model} uses all epochs to train one model.
}

  \small
  \begin{tabular}{|c|c|c|c|c|c|c|}
    \hline  Task		& Strategy		& Min		& Max		& Median
    & Mean		& Std \\
    \hline\hline  \multirow{3}{3cm}{\centering Sine problem\\ test squared loss}		& student		& \textbf{0.049}		& \textbf{0.116}		& \textbf{0.078}		& \textbf{0.077}		& \textbf{0.015} \\
    \cline{2-7}  & bo3 model		& 0.105		& 0.373		& 0.253		& 0.240		& 0.076 \\
    \cline{2-7}  & one model		& 0.053		& 0.363		& 0.281		& 0.249		& 0.097 \\
    \hline
  \end{tabular}

\end{table*}

\begin{figure}
    \centering
  
    \includegraphics[width=0.7\textwidth,clip,trim=0 9cm 7cm 0]{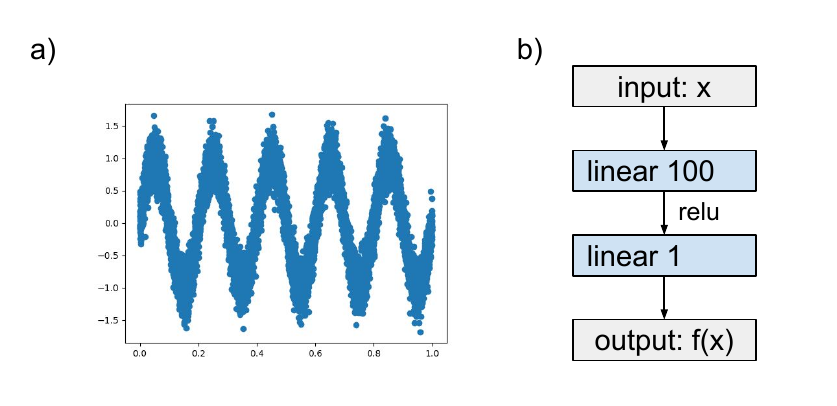}
    
    \caption{
       \label{fig:sine} a) Training dataset for sine problem consisting of 10000 samples where $x \sim
       \mathcal{U}(0,1)$ and $y = sin(10\pi x) + z \, ;
         \: z \sim \mathcal{N}(0,\,0.2)$.
        b) Architecture of model for sine problem
    }
  \end{figure}

First, we want to confirm the idea that a network trained from random initialization might end in suboptimal local optima and our merging procedure finds higher quality local optima. 
To verify, we have created a synthetic dataset - five sine waves with noise.
The input is scalar $x$ and the target is $y = sin(10\pi x)+z$ where $z \sim \mathcal{N}(0,\,0.2)$, see Fig.
  \ref{fig:sine}.

Our architecture is composed of two linear layers (i. e. one hidden layer) with 100 hidden neurons
(Fig. \ref{fig:sine}).
In every strategy, we have used 900 epochs and SGD with starting learning rate 0.01 and momentum 0.9.
Then we have decreased the learning rate to 0.001 after the 100th epoch for student fine-tuning, the 250th epoch for teachers and bo3 models, and the 800th epoch for a model in \textit{one model}.  
We repeat all experiments 50 times. 

\begin{figure}[t!]
    \centering
  
    \includegraphics[width=0.6\textwidth]{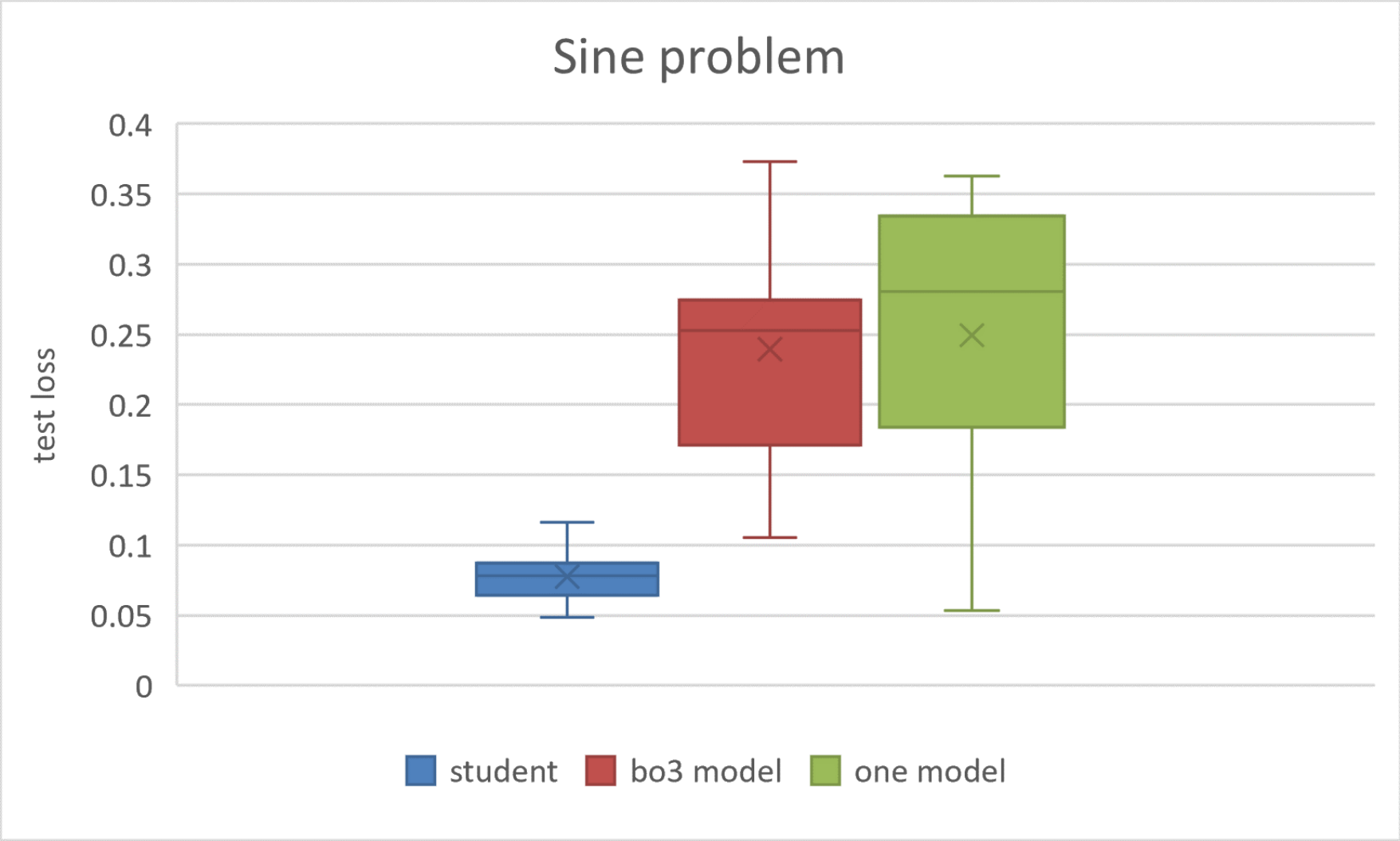}
    
    \caption{
       \label{fig:sine_box_plot} Box plot of testing losses of 50 experiments on sine problem.
        The vertical line inside the box represents the median, and the cross presents the mean.
    }
  \end{figure}

We can observe, that our strategy has significantly smaller error than other strategies (Table.
~\ref{fig:sine_results_tables}, Fig.~\ref{fig:sine_box_plot}).

\begin{figure}
    \centering
  
    \includegraphics[width=0.9\textwidth]{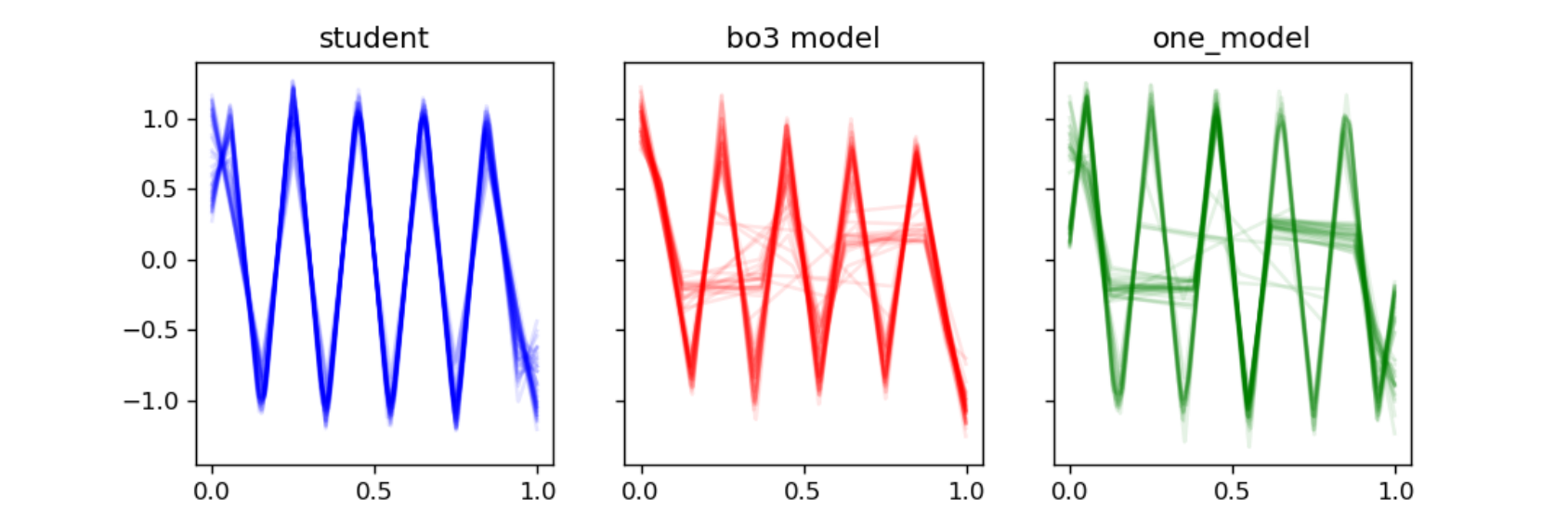}
    
    \caption{
      \label{fig:learned_curve} Plots of learned sine curves by models trained with different strategies.
          We plot every train result as one line and overlay on top of each other.
          As we can see, all of the students resulting from merging get all of the peaks.
          However, models without merging often missed some peaks.}
    
  \end{figure}

Digging deeper (Fig.~\ref{fig:learned_curve}), we observe networks trained by our strategy to predict all the peaks correctly. 
However, networks trained by generic strategy often miss some peaks. This indicates that our training strategy helps the network to select better features for later use.
In some cases, bo1 network is lucky and predicts all the peaks correctly, which can be seen in having
similar minimal error as our training strategy.

\subsection{\textbf{Image classification}}

Here we test our training strategy on various combinations of dataset and architecture.
First, we use Imagewoof (Imagenet-1k using only 10 classes of dog breeds) dataset \cite{imagewoof,shleifer2019using} with LeNet
\cite{lecun1989handwritten} and ResNet18 \cite{he2016deep} architectures.
Then, we test our approach on CIFAR-100 dataset \cite{krizhevsky2009learning} using ResNet20
\cite{he2016deep} architecture.
Finally, we evaluate our approach on Imagenet-1k dataset \cite{deng2009large}.

In all cases, our training strategy with merging provides better results than generic training strategies. 
Results are sumarized in Tables \ref{fig:statistics_of_all_experiments} and \ref{fig:imagenet} and
details about training setup are provided below.

\begin{table*}[t!]
  \centering

    \caption{
 \label{fig:statistics_of_all_experiments} Summary of experimental results on image classification tasks.
 We report testing accuracy.
 Considering specific task, all strategies used the same number of epochs.
  Strategy \textit{student} (our strategy) uses 2/3 to train two teachers and 1/3 to train student (1/6
  finding important features, 1/6 finetuning).
  Strategy \textit{bo3 model} trains three models and picks the best.
  Strategy \textit{one model} uses all epochs to train one model.
}
  \begin{tabular}{|c|c|c|c|c|c|c|}
    \hline  Task		& Strategy		& Min		& Max		& Median
    & Mean		& Std \\
    \hline\hline  \multirow{3}{3.5cm}{\centering Imagewoof \\ LeNet \\ test acc [\%]}		&
    student		& \textbf{38.0}		& \textbf{41.1}		& \textbf{39.9}		&
    \textbf{39.9}		& 0.8 \\
    \cline{2-7}  &bo3 model		& 36.9		& 39.4		& 38.7		& 38.5
    & \textbf{0.7} \\
    \cline{2-7}  &one model		& 36.5		& 39.7		& 37.7		& 37.8
    & 1.0 \\
    \hline\hline  \multirow{3}{3.5cm}{\centering Imagewoof \\ ResNet18 \\ test acc [\%]}		&
    student		& \textbf{82.4}		& \textbf{82.8}		& \textbf{82.5}		&
    \textbf{82.6}		& \textbf{0.2} \\
    \cline{2-7}  &bo3 model		& 80.1		& 80.9		& 80.7		& 80.6		& 0.3 \\
    \cline{2-7}  &one model		& 81.0		& 81.8		& 81.3		& 81.3		& 0.3 \\

    \hline\hline  \multirow{3}{3.5cm}{\centering CIFAR-100 \\ ResNet20 \\ test acc [\%]}		&
    student		& \textbf{68.8}		& \textbf{69.1}		&
    \textbf{68.8}		& \textbf{68.9}		& 0.2 \\
    \cline{2-7}  &bo3 model		& 67.0 	& 67.2		& 67.0		& 67.0
    & \textbf{0.1} \\
    \cline{2-7}  &one model		& 67.0		& 67.9        & 67.5	& 67.5		&   0.4 \\
    \hline
  \end{tabular}

\end{table*}

\subsubsection{\textbf{Imagewoof on LeNet}}

LeNet is composed of two convolutional layers followed by three linear layers.
  The shape of an input image is (28, 28, 3).
  The convolutional layers have 6 and 16 output channels, respectively.
  The linear layers have 400, 120, and 80 input features, respectively.
   For the architecture of the big student see Fig.~\ref{fig:network_with_gates}.

Every strategy has used 6000 epochs cumulatively and SGD with starting learning rate 0.01 and momentum 0.9.
  Every training except finding important neurons (teachers, student fine-tuning, bo3 models, and one model) decreased the learning to 0.001 in the third quarter and 0.0001 in the last quarter of the training.

\begin{figure}[t!]
    \centering
  
    \includegraphics[width=0.6\textwidth]{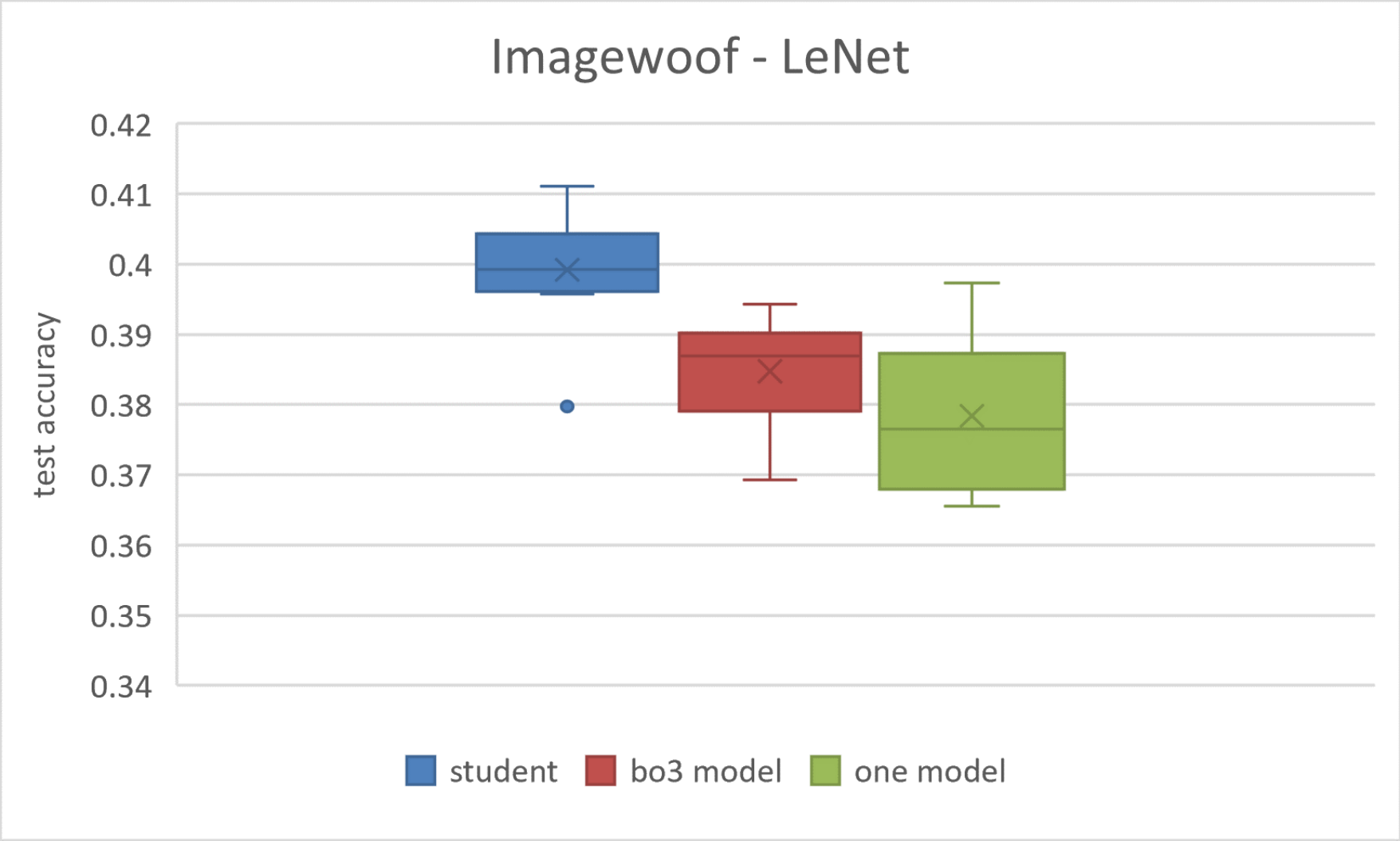}
    
    \caption{
       \label{fig:imagewoof_lenet} Box plot of the testing accuracies of 10 experiments on Imagewoof
       with LeNet.
    }
  \end{figure}

We have conducted 10 experiments, see Fig.~\ref{fig:imagewoof_lenet} for visualisation and Table \ref{fig:statistics_of_all_experiments} for detailed statistics.
  Our strategy has consistently better results than other strategies.
  It has a greater sample variance than \textit{one model} as a consequence of an outlier, see Fig.~\ref{fig:imagewoof_lenet}.

\subsubsection{\textbf{Imagewoof on ResNet18}}

ResNet has two information flows (one through blocks, one through skip connections).
  Throughout the computation, its update is $x = f(x) + x$, instead of the original $x = f(x)$.
  To conserve this property, some gate layers have to be synchronized - share weights and realization of random variables, see \ref{fig:network_with_gates}.

Every strategy has used 600 epochs cumulatively.
  The optimizer and the learning rate scheduler is analogical to the LeNet experiment.

\begin{figure}
    \centering
  
    \includegraphics[width=0.6\textwidth]{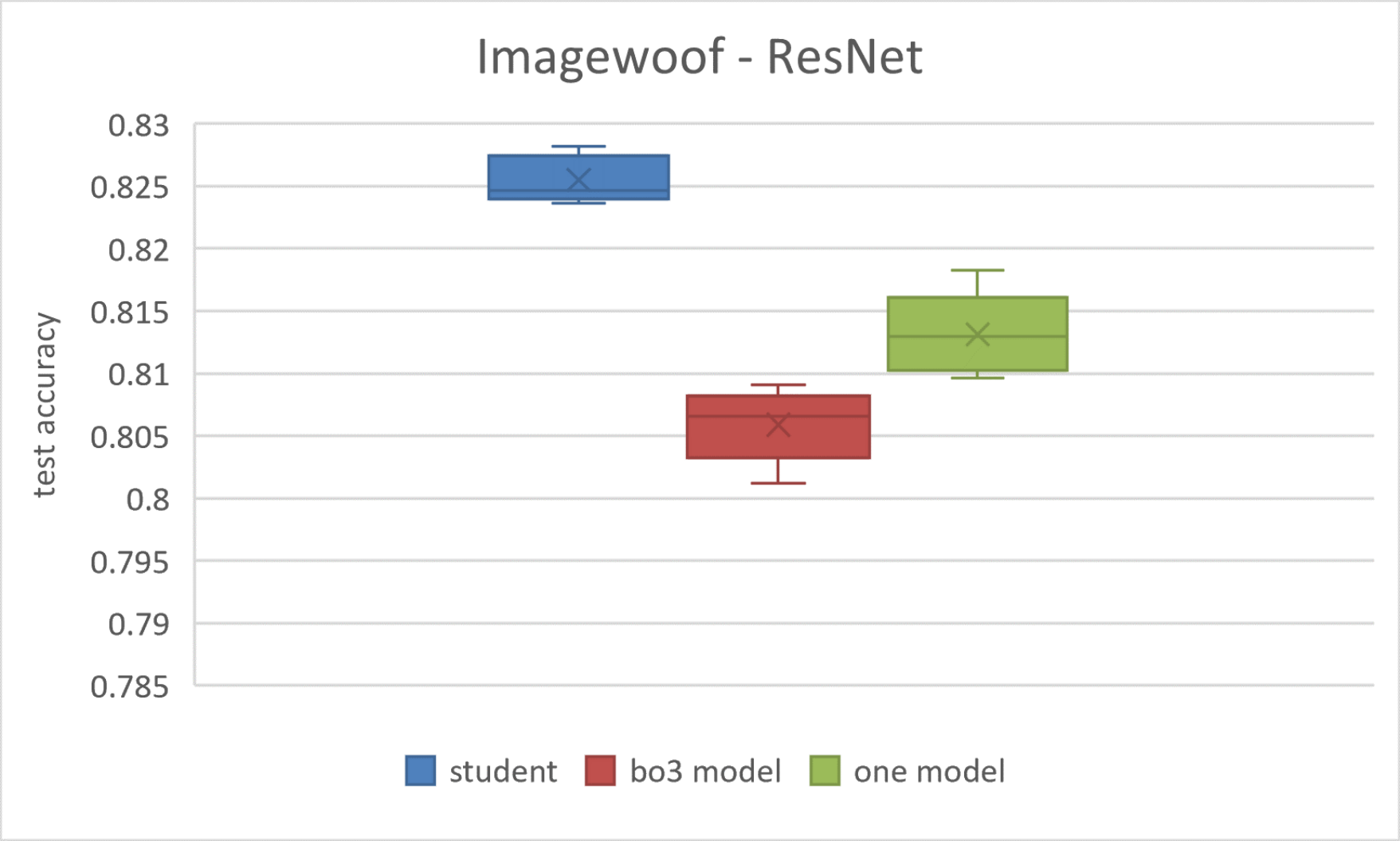}
    
    \caption{
      \label{fig:imagewoof_resnet} Results of 5 experiments on Imagewoof with ResNet18.
          The worst student (0.819) had slightly better accuracy than the best long teacher (0.818).}
    
  \end{figure}

We have conducted 5 experiments, see Fig.~\ref{fig:imagewoof_resnet} for visualisation and table \ref{fig:statistics_of_all_experiments} for detailed statistics.
Similarly, as with LeNet, our strategy has consistently better results than other strategies.

\subsubsection{\textbf{CIFAR-100 on ResNet20}}

We also tested our approach on CIFAR-100 dataset using ResNet20.
Our total training budget is 900 epochs. We optimize models using SGD with starting learning rate
0.1 and then divide it by 10 during half and three quarters of the training of one network.
We run all strategies 5 times and report results in Table \ref{fig:statistics_of_all_experiments}.
We can see, that our strategy is more than $1\%$ better than training one model for extended period
of time.

\subsubsection{Imagenet on ResNet18}

\begin{table*}
\centering
\caption{\label{fig:imagenet} Results on ResNet-18 on Imagenet benchmark.}

\begin{tabular}{|c|c|c|c|c|}
\hline
Teacher & Big student & Finetuning & Total & Validation \\
epochs  & epochs      & epochs     & epochs & accuracy  \\\hline
-              & -                  & 90                & 90           & 0.6976 \\\hline
-              & -                  & 150                & 150           & 0.7028 \\\hline
20             & 20                 & 90                 & 150           & {\bf 0.7047} \\\hline

\end{tabular}
\end{table*}

We tested our merging approach also on Imagenet-1k dataset \cite{deng2009large}.
However as seen in \cite{wightman2021resnet} high quality training requires 300 to 600 epoch, which is quite prohibitive. 
We opted for approach from Torchvision \cite{Torchvision}, which achives decent results in 90
epochs. We train networks using SGD with starting learning rate 0.1, which decreases by factor of 10
in third and two thirds of training. For final finetuning of student we used slightly smaller starting learning rate of $0.07$.

For merging we used slightly different appoach than in previous experiments.
We trained teachers only for short amount of 20 epochs, which gives teacher accuracy around $65\%$.
Then we spend 20 epochs in tuning big student and finding important neurons and finally finetune for
90 epochs. In total of 150 epochs, we get better results than ordinary training for 90 epochs and also better result than training for equivalent amount of 150 epochs.
Results are summarized in Tab. \ref{fig:imagenet}.

\section{{Conclusions and future work}}

We proposed a simple scheme for merging two neural networks trained from different initializations into one.
Our scheme can be used as a finalization step after one trains multiple copies of one network with varying starting seeds.
Alternatively, we can use our scheme for getting higher quality networks under a similar training budget which we experimentally demonstrated.

One of the downsides of our scheme is that during selection of important neurons we need to instantiate rather big neural network.
In future, we would like to optimize this step to be more resource efficient. One option is to select important neurons in layerwise fashion.
  
Other possible options for future research include merging more than two networks and also merging networks pretrained on different datasets.

\section{{Acknowledgements}}

This research was supported by grant 1/0538/22 from Slovak research grant agency VEGA.

%
%
%
\bibliographystyle{elsarticle-num}
\bibliography{main}

\begin{thebibliography}{10}
\expandafter\ifx\csname url\endcsname\relax
  \def\url#1{\texttt{#1}}\fi
\expandafter\ifx\csname urlprefix\endcsname\relax\def\urlprefix{URL }\fi
\expandafter\ifx\csname href\endcsname\relax
  \def\href#1#2{#2} \def\path#1{#1}\fi

\bibitem{picard2021torch}
D.~Picard, Torch. manual\_seed (3407) is all you need: On the influence of
  random seeds in deep learning architectures for computer vision, arXiv
  preprint arXiv:2109.08203 (2021).

\bibitem{wightman2021resnet}
R.~Wightman, H.~Touvron, H.~J{\'e}gou, Resnet strikes back: An improved
  training procedure in timm, arXiv preprint arXiv:2110.00476 (2021).

\bibitem{molchanov2016pruning}
P.~Molchanov, S.~Tyree, T.~Karras, T.~Aila, J.~Kautz, Pruning convolutional
  neural networks for resource efficient inference, arXiv preprint
  arXiv:1611.06440 (2016).

\bibitem{molchanov2019importance}
P.~Molchanov, A.~Mallya, S.~Tyree, I.~Frosio, J.~Kautz, Importance estimation
  for neural network pruning, in: Proceedings of the IEEE/CVF Conference on
  Computer Vision and Pattern Recognition, 2019, pp. 11264--11272.

\bibitem{luo2017thinet}
J.-H. Luo, J.~Wu, W.~Lin, Thinet: A filter level pruning method for deep neural
  network compression, in: Proceedings of the IEEE international conference on
  computer vision, 2017, pp. 5058--5066.

\bibitem{yu2018nisp}
R.~Yu, A.~Li, C.-F. Chen, J.-H. Lai, V.~I. Morariu, X.~Han, M.~Gao, C.-Y. Lin,
  L.~S. Davis, Nisp: Pruning networks using neuron importance score
  propagation, in: Proceedings of the IEEE Conference on Computer Vision and
  Pattern Recognition, 2018, pp. 9194--9203.

\bibitem{hinton2015distilling}
G.~Hinton, O.~Vinyals, J.~Dean, et~al., Distilling the knowledge in a neural
  network, arXiv preprint arXiv:1503.02531 2~(7) (2015).

\bibitem{beyer2021knowledge}
L.~Beyer, X.~Zhai, A.~Royer, L.~Markeeva, R.~Anil, A.~Kolesnikov, Knowledge
  distillation: A good teacher is patient and consistent, arXiv preprint
  arXiv:2106.05237 (2021).

\bibitem{nath2020better}
U.~Nath, S.~Kushagra, Better together: Resnet-50 accuracy with 13x fewer
  parameters and at 3x speed, arXiv preprint arXiv:2006.05624 (2020).

\bibitem{chen2022the}
T.~Chen, Z.~Zhang, Y.~Cheng, A.~H. Awadallah, Z.~Wang, The principle of
  diversity: Training stronger vision transformers calls for reducing all
  levels of redundancy, in: In IEEE Conference on Computer Vision and Pattern
  Recognition (CVPR 2022), 2022.

\bibitem{voita2019analyzing}
E.~Voita, D.~Talbot, F.~Moiseev, R.~Sennrich, I.~Titov, Analyzing multi-head
  self-attention: Specialized heads do the heavy lifting, the rest can be
  pruned, arXiv preprint arXiv:1905.09418 (2019).

\bibitem{louizos2017learning}
C.~Louizos, M.~Welling, D.~P. Kingma, Learning sparse neural networks through $
  l\_0 $ regularization, arXiv preprint arXiv:1712.01312 (2017).

\bibitem{imagewoof}
J.~Howard, Imagenette, \url{https://github.com/fastai/imagenette}.

\bibitem{shleifer2019using}
S.~Shleifer, E.~Prokop, Using small proxy datasets to accelerate hyperparameter
  search, arXiv preprint arXiv:1906.04887 (2019).

\bibitem{lecun1989handwritten}
Y.~LeCun, B.~Boser, J.~Denker, D.~Henderson, R.~Howard, W.~Hubbard, L.~Jackel,
  Handwritten digit recognition with a back-propagation network, Advances in
  neural information processing systems 2 (1989).

\bibitem{he2016deep}
K.~He, X.~Zhang, S.~Ren, J.~Sun, Deep residual learning for image recognition,
  in: Proceedings of the IEEE conference on computer vision and pattern
  recognition, 2016, pp. 770--778.

\bibitem{krizhevsky2009learning}
A.~Krizhevsky, G.~Hinton, et~al., Learning multiple layers of features from
  tiny images (2009).

\bibitem{deng2009large}
J.~Deng, A large-scale hierarchical image database, Proc. of IEEE Computer
  Vision and Pattern Recognition, 2009 (2009).

\bibitem{Torchvision}
Torchvision, \url{https://pytorch.org/vision/stable/index.html}.

\end{thebibliography}
\end{document}